\documentclass[manuscript,screen]{acmart}

\AtBeginDocument{%
  }

\setcopyright{acmlicensed}
\acmJournal{THRI}
\acmYear{2025}
\copyrightyear{2025}

\usepackage{textcomp}
\usepackage{booktabs}
\usepackage{threeparttable}
\usepackage{pifont}
\usepackage[table]{xcolor}
\usepackage{adjustbox}
\usepackage{multirow}
\usepackage{graphicx}
\usepackage{subcaption}
\usepackage{soul, color}
\definecolor{Gray}{gray}{0.9}
\newcommand{\xmark}{\ding{55}}
\renewcommand{\hl}[1]{#1}

\title{DiG-Net: Enhancing Human–Robot Interaction through Hyper-Range Dynamic Gesture Recognition in Assistive Robotics}

\author{Eran Bamani Beeri}
\email{eran0910@mit.edu}
\affiliation{%
  \institution{Massachusetts Institute of Technology (MIT)}
  \city{Cambridge}
  \state{Massachusetts}
  \country{USA}
}

\author{Eden Nissinman}
\affiliation{%
  \institution{Tel Aviv University}
  \city{Tel Aviv}
  \country{Israel}
}

\author{Avishai Sintov}
\affiliation{%
  \institution{Tel Aviv University}
  \city{Tel Aviv}
  \country{Israel}
}

\renewcommand{\shortauthors}{Beeri et al.}

\begin{abstract}

Dynamic hand gestures play a pivotal role in assistive human-robot interaction (HRI), facilitating intuitive, non-verbal communication, particularly for individuals with mobility constraints or those operating robots remotely. Current gesture recognition methods are mostly limited to short-range interactions, reducing their utility in scenarios demanding robust assistive communication from afar. \hl{In this paper, we present DiG-Net, the first dynamic gesture recognition framework enabling robust operation at hyper-range distances of up to 30 meters, specifically designed for assistive robotics to enhance accessibility and improve quality of life.} Our proposed Distance-aware Gesture Network (DiG-Net) effectively combines Depth-Conditioned Deformable Alignment (DADA) blocks with Spatio-Temporal Graph modules, enabling robust processing and classification of gesture sequences captured under challenging conditions, including significant physical attenuation, reduced resolution, and dynamic gesture variations commonly experienced in real-world assistive environments. We further introduce the Radiometric Spatio‐Temporal Depth Attenuation Loss (RSTDAL), shown to enhance learning and strengthen model robustness across varying distances. Our model demonstrates significant performance improvement over state-of-the-art gesture recognition frameworks, achieving a recognition accuracy of 97.3\% on a diverse dataset with challenging hyper-range gestures. By effectively interpreting gestures from considerable distances, DiG-Net significantly enhances the usability of assistive robots in home healthcare, industrial safety, and remote assistance scenarios, enabling seamless and intuitive interactions for users regardless of physical limitations.

\end{abstract}



\begin{document}

\maketitle

\section{Introduction}

The growing number of individuals living with disabilities and requiring assistance has created a pressing demand for assistive technologies that enhance users’ independence, safety, and quality of life~\cite{nanavati2023physically}. Among these, assistive robotic systems are increasingly integrated into environments where intuitive, nonverbal communication is essential for enabling natural interaction with individuals of varied abilities. Gesture-based interaction is particularly important in scenarios where speech is not an option. To contextualize our contribution, Table~\ref{tab:gesture_assistive_comparison} presents a comparative overview of recent systems in this area, outlining their target users, sensing modalities, application domains, and level of human involvement. \hl{Recent developments in gesture-based assistive systems have improved short-range communication, yet reliable operation at long distances remains largely unexplored. This work addresses that gap by enabling robust dynamic gesture recognition suitable for real-world assistive deployment.} Recent research has catalyzed a new generation of assistive systems capable of perceiving complex environments and interacting with humans in context-aware, natural ways~\cite{leal2019computer, robinson2023robotic, drolshagen2023context}. Equally important, human-in-the-loop approaches are being developed to ensure these systems operate safely and reliably alongside people. These methods incorporate human feedback, predictive planning, and adaptive control to support trust and safety during close interactions~\cite{lasota2017survey, safeea2019minimum, losey2018review}. To enable practical deployment, recent studies stress the importance of user-centered design and usability evaluation, ensuring assistive robots are intuitive, reliable, and effective~\cite{frennert2024critical, shourmasti2021user}. A key area of research supporting these goals is HRI, which focuses on enabling smooth, natural collaboration between people and robots, both in assistive contexts and everyday life~\cite{goodrich2008human, maurtua2017natural}. This reflects a broader shift from viewing robots as tools to recognizing them as partners that augment and support human capabilities~\cite{mataric2017socially}.

Beyond these technological advances, recent literature in HRI emphasizes that social perception, trust, and intuitiveness are equally essential for effective collaboration. 
Nonverbal cues, such as gestures, posture, and motion timing, play a fundamental role in how humans interpret robot behavior and establish mutual understanding. 
Urakami and Seaborn~\cite{urakami2023nonverbal} highlighted that natural gesture expressiveness and responsiveness significantly affect user comfort and perceived agency during interaction. 
Similarly, Robinson et al.~\cite{robinson2023robotic} stressed the growing importance of robotic vision that supports collaboration through real-time, context-aware perception. 
As Mataric~\cite{mataric2017socially} argued in her influential review on socially assistive robotics, the ultimate goal is to move from automation toward human augmentation, developing robots that enhance human capability rather than replace it. 
Framing DiG-Net within this paradigm positions it not merely as a gesture recognition model, 
but as a medium for intuitive, trustworthy, and socially aware communication between humans and assistive robots.

\begin{table}[htbp]
\centering
\small
\setlength{\tabcolsep}{4pt}
\caption{Comparative summary of gesture-based assistive robotics systems.}
\label{tab:gesture_assistive_comparison}
\begin{threeparttable}
\begin{adjustbox}{width=\linewidth}
\begin{tabular}{l l c c l c c}
\toprule
\textbf{Paper} & \textbf{Population} & \textbf{Env.} & \textbf{Modality} & \textbf{Application} & \textbf{Range} & \textbf{Human-in-the-loop} \\
\midrule
Boboc et al. \cite{boboc2015point}   & General users          & Indoor & RGB-D & Home robot control        & Short & Yes \\
Haseeb et al. \cite{haseeb2018head}  & Motor-impaired         & Indoor & IMU   & Head-based robot control  & N/A   & Yes \\
Rudigkeit et al. \cite{rudigkeit2019amicus} & Tetraplegic        & Indoor & IMU   & Robot arm use             & N/A   & Yes \\
Yang et al. \cite{yang2025high}      & Upper-limb impaired    & Indoor & EMG   & Mobile manipulation       & Long  & Yes \\
Ababneh et al. \cite{ababneh2018gesture} & Elderly, wheelchair  & Indoor & RGB-D & Arm control via gesture   & Short & Yes \\
Neto et al. \cite{neto2019gesture}   & Factory workers        & Indoor & IMU   & Assembly support          & Mid   & Partial \\
Werner et al. \cite{werner2020improving} & Elderly             & Indoor & RGB-D & Bathing assistant         & Short & Partial \\
Muñoz et al. \cite{munoz2021kinect}  & Older adults           & Indoor & RGB-D & Rehab \& exercises        & Short & Partial \\
Oudah et al. \cite{oudah2020elderly} & Speech-impaired elderly & Indoor & RGB-D & Emergency alerting       & Mid   & Yes \\
Bandara et al. \cite{bandara2020intelligent} & Wheelchair users  & Indoor & RGB-D & Wheelchair navigation     & Short & Yes \\
\midrule
\textbf{This work} & \textbf{General users} & \textbf{Both} & \textbf{RGB} & \textbf{Remote HRI guidance} & \textbf{Hyper} & \textbf{Yes} \\
\bottomrule
\end{tabular}
\end{adjustbox}
\end{threeparttable}
\end{table}

Intuitive interaction mechanisms are essential to enable non-expert users to effectively communicate their intentions to robots. 
Hand and arm gestures are intuitive forms of nonverbal communication, widely used in daily human interactions, making them well-suited for HRI \cite{gao2020robust, wang2022hand, urakami2023nonverbal}. Gesture-based interactions reduce the need for complex verbal or physical interfaces, improve user experience, and allow for intuitive and rapid communication with robots, even for users with no technical expertise \cite{wachs2011vision}. With gestures, a user can convey nonverbal and simple commands even from a long distance without the need to shout. For instance, a user may direct robot movements with pointing gestures \cite{bamani2025real}.

\begin{figure}
    \centering
    \includegraphics[width=0.7\linewidth]{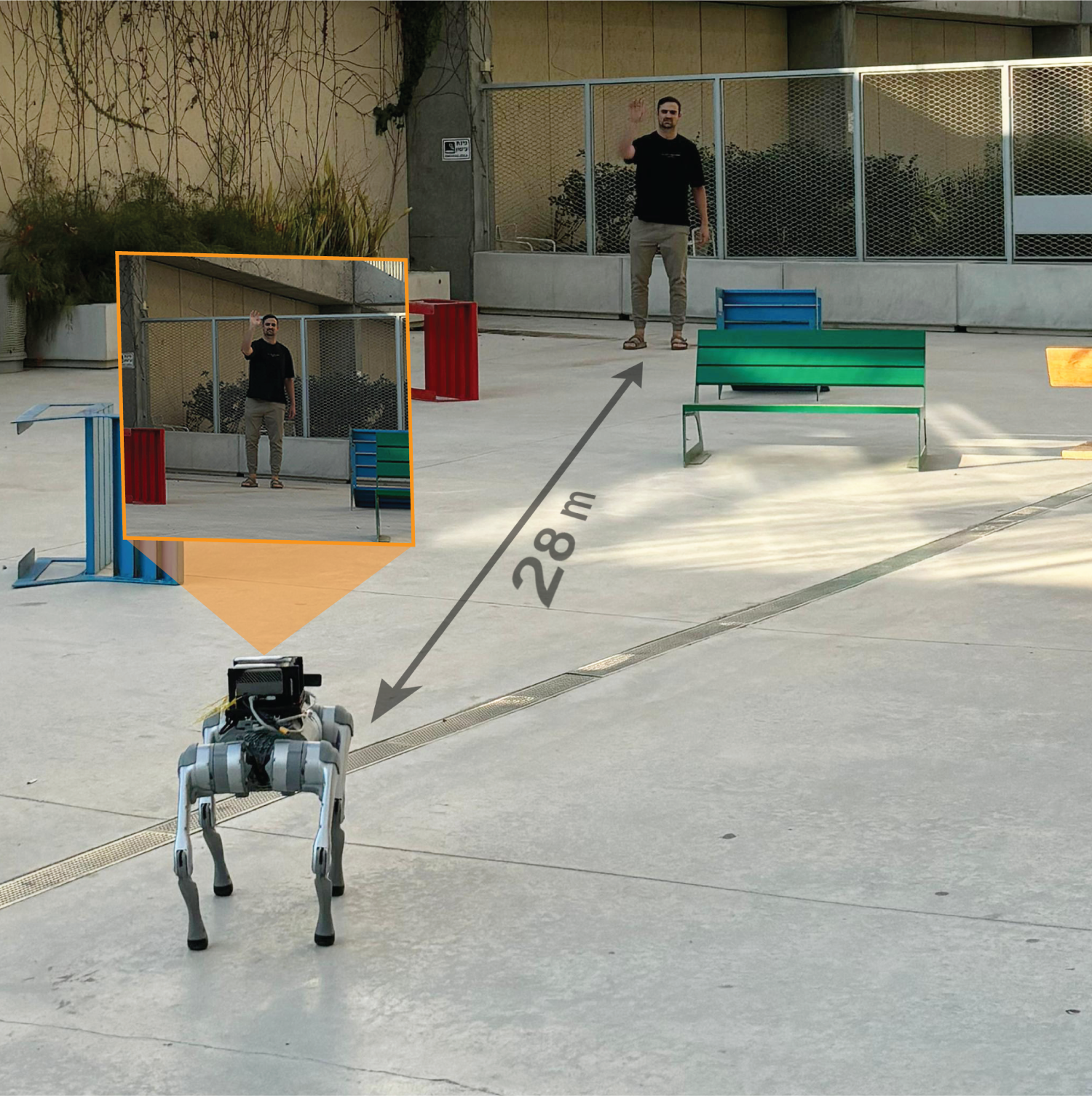}
    \caption{Demonstration of a user instructing a robot to go back by sweeping an open palm forward and backward, from a hyper-range distance. In addition to the low-resolution view of the user's hand, the robot may confuse the dynamic gesture with the static stop gesture.}
    \Description{A person standing far from a mobile robot performs a dynamic back-and-forth palm gesture. 
    The image shows how low resolution at a long distance can cause the robot to misinterpret the motion as a static stop signal.}
    \label{fig:Front}
\end{figure}

Research in the field of gesture recognition has made significant progress, particularly in recognizing static gestures \cite{zhang2020racon, yu2021multi,Yu2022}. However, most gesture recognition approaches are limited to short-range interactions, typically within a few meters \cite{nickel2007visual, wachs2011vision, wang2022hand}. Moreover, gesture recognition methods, often described to be effective in the long range, are typically limited to around seven meters \cite{Zhou9561189,Liang2024}. However, this limitation hinders their applicability in real-world scenarios that demand longer-range HRI. Recognizing gestures from a truly long distance can significantly expand the potential applications of robots in environments such as public spaces, industrial settings, and emergencies, where natural, non-contact interactions are necessary \cite{kim2014non, canal2015gesture}. However, one of the key challenges in achieving effective long-range gesture recognition is the degradation of visual information due to factors such as reduced resolution, lighting variations, and occlusions \cite{shen2019overview, pla2018three}. 

While our recent work \cite{bamani2024ultra} successfully demonstrated ultra-range recognition of static hand gestures using a simple web camera at distances up to 25 meters, recognizing dynamic gestures presents additional, significant challenges. \hl{Static gesture recognition relies primarily on spatial features captured in a single frame, whereas dynamic gesture recognition demands the extraction and analysis of temporal sequences across multiple frames. Prior works have addressed challenges such as occlusion and dynamic ambiguity in short-range settings, for example, by using shape-based descriptors or customized tracking to handle missing or discontinuous cues}~\cite{xing2019dynamic, han2018robust, jain2023linked}. \hl{However, these approaches typically assume that the hand region is sufficiently resolved and that reliable hand-configuration cues are available, assumptions that can break down at long distances where the gesture occupies only a small portion of the image. As a result, methods designed for short-range sensing may not transfer directly to long-distance, RGB-only deployment without explicitly accounting for compounded degradations.} Dynamic gesture recognition at hyper-range presents a unique challenge that demands the fusion of spatial and temporal information across multiple scales and distances. At such long ranges, the visual signal of the performer is severely diminished, the region of interest occupies only a small portion of the frame, resulting in low-resolution observations, and environmental noise (e.g., background clutter, illumination changes, atmospheric effects) can significantly degrade each frame. In this scenario, temporal cues become as crucial as spatial cues for reliable recognition. Fast, subtle hand motions (e.g., a quick finger wave) might be nearly imperceptible in any single low-resolution frame, yet can be captured by analyzing differences across consecutive frames. Conversely, the overall spatial configuration of a gesture (e.g., an arm extended forward) may be visible but ambiguous without a temporal context to distinguish its meaning. This interplay between space and time at extreme distances requires an architecture that can simultaneously capture fine, rapid movements and aggregate information over longer temporal intervals. 


\begin{figure}[htbp]
    \centering
    \includegraphics[width=\linewidth]{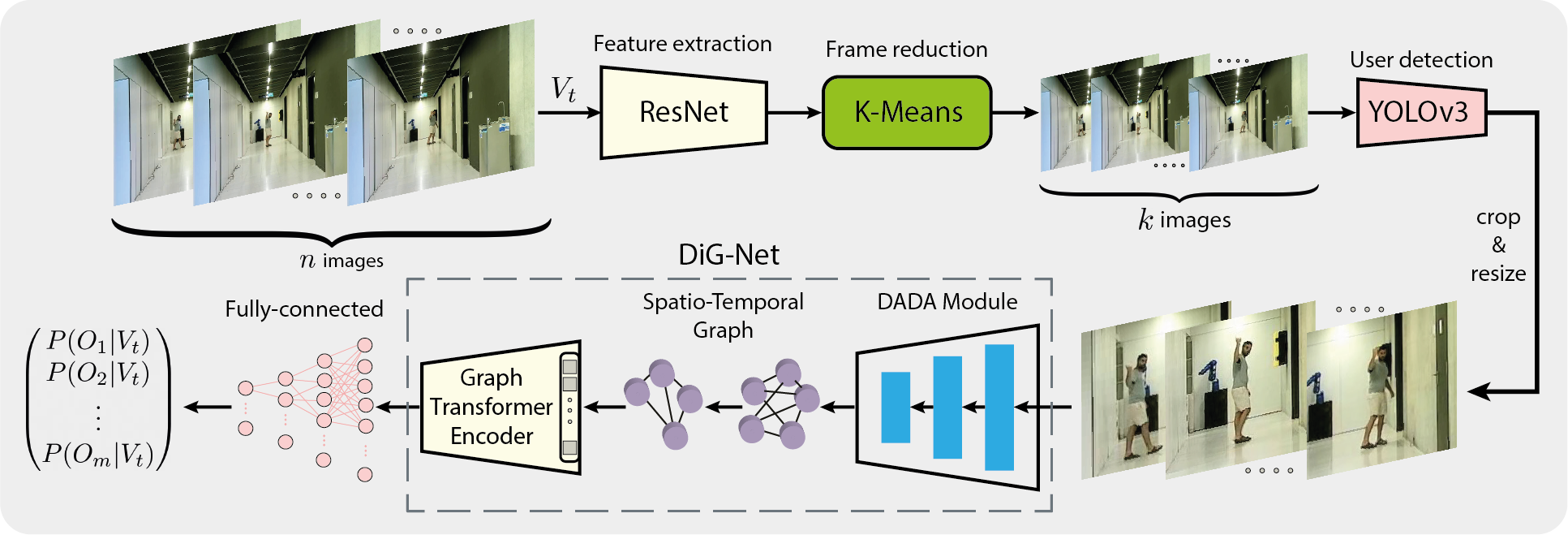}
    \caption{Overview of the proposed DiG-Net framework for hyper-range dynamic hand gesture recognition.
    The model combines Depth-Conditioned Deformable Alignment (DADA), Spatio-Temporal Graph (STG) modules, and Graph Transformer encoders to recognize gestures from RGB videos at distances up to 30 meters.}
    \Description{A schematic diagram illustrating the DiG-Net architecture for hyper-range dynamic gesture recognition.
    The pipeline begins with feature extraction via ResNet, followed by K-means-based frame reduction and YOLOv3 user detection.
    Processed frames are fed into the DiG-Net, which integrates DADA blocks for depth-conditioned alignment, STG modules for temporal reasoning, and Graph Transformer encoders for spatial-temporal fusion, ending with a classification head outputting gesture classes.}
    \label{fig:scheme}
\end{figure}

Dynamic gestures provide valuable context for understanding actions and systems, especially when explaining processes that unfold over time \cite{Kang2016}. They are particularly effective for conveying spatial context to a task command \cite{Clough2020}. For instance, a gesture instructing a robot to move backward would usually include an open palm swept back and forth toward it, as demonstrated in Figure \ref{fig:Front}. However, a snapshot of the gesture may be recognized as a static stop instruction if not processed correctly. Several studies have attempted to address dynamic gesture recognition by utilizing various data modalities. RGB-Depth (RGB-D) cameras are often used to recognize dynamic gestures but are limited to indoor environments \cite{xu2015online,ma2018kinect,kabir2019novel, bokstaller2021dynamic,Gao2022}. Approaches using simple RGB cameras were also proposed while shown to function only indoors \cite{Zhou9561189,Yi2018,dos2020dynamic}. Table \ref{tb:sota} summarizes the prominent work on dynamic gesture recognition with visual perception, highlighting the limitations of current approaches, which are typically restricted to indoor environments and short-range interactions (within 7 meters). In it also worth noting on-body sensors in wearable devices that instantly recognize hand gestures \cite{Pyun2023, Tchantchane2023}. While they can offer high recognition rates, wearable devices often require specialized and expensive hardware, limiting their accessibility to occasional users. Additionally, biometric approaches may not generalize well to new users, requiring additional data collection for each individual. On the other hand, visual perception allows for the observation of any user.


\begin{table}[htbp]
\centering
\small
\setlength{\tabcolsep}{4pt}
\caption{State-of-the-art comparison of dynamic gesture recognition methods using visual perception.}
\label{tb:sota}
\begin{threeparttable}
\begin{adjustbox}{width=\linewidth}
\begin{tabular}{l c c c c}
\toprule
\textbf{Paper} & \textbf{Camera} & \textbf{Range} & \textbf{Indoor} & \textbf{Outdoor} \\
\midrule
Zhou et al. \cite{Zhou9561189} & RGB            & $\leq$7\,m  & \checkmark & \xmark \\
Xu et al. \cite{xu2015online}  & RGB-D          & $<$1\,m     & \checkmark & \xmark \\
Ma et al. \cite{ma2018kinect}  & RGB-D          & $\leq$2\,m  & \checkmark & \xmark \\
Kabir et al. \cite{kabir2019novel} & RGB-D       & $<$1\,m     & \checkmark & \xmark \\
Bokstaller et al. \cite{bokstaller2021dynamic} & RGB-D & – & \checkmark & \xmark \\
Gao et al. \cite{Gao2022}      & RGB-D          & $<$2\,m     & \checkmark & \xmark \\
Yi et al. \cite{Yi2018}        & RGB            & $\leq$6\,m  & \checkmark & \xmark \\
dos Santos et al. \cite{dos2020dynamic} & RGB    & –           & \checkmark & \xmark \\
Tran et al. \cite{tran2019dynamic} & 5$\times$RGB-D & $<$2\,m   & \checkmark & \xmark \\
Wu et al. \cite{wu2016deep}    & RGB-D          & $<$2\,m     & \checkmark & \xmark \\
\midrule
\textbf{Proposed method (DiG-Net)} & \textbf{RGB} & \textbf{$\leq$30\,m} & \textbf{\checkmark} & \textbf{\checkmark} \\
\bottomrule
\end{tabular}
\end{adjustbox}
\end{threeparttable}
\end{table}

While the above methods show feasibility in controlled indoor settings, they often lack scalability and robustness for outdoor environments or hyper-range interactions. Their reliance on specialized hardware, such as depth cameras, can increase cost and complexity, limiting their practical applicability \cite{ji20123d}. Recent approaches utilizing multimodal data, including RGB, depth, and human-pose (i.e., skeleton) information, have demonstrated potential for dynamic gesture recognition \cite{tran2019dynamic, wu2016deep}. However, these methods are often also constrained by limited operational range and the need for specialized equipment. This makes them less suitable for scenarios requiring long-range interactions or deployment in large, open spaces. To the best of the authors' knowledge, no work has addressed the problem of recognizing dynamic gestures at distances farther than seven meters.

In addition to capturing temporal features of dynamic gestures, the challenge is further compounded by the difficulty of perceiving spatial motions at far distances. In this paper, we address the problem of dynamic gesture recognition at hyper-range distance by only using a simple RGB camera. Specifically, we propose the Distance-aware Gesture Network (DiG-Net), illustrated in Figure \ref{fig:scheme}, which comprises a novel Depth-Conditioned Deformable Alignment (DADA) block, seamlessly fused with Spatio-Temporal Graph (STG) modules \cite{yan2018spatial}. Unlike existing approaches, DiG-Net adaptively warps feature maps based on per-pixel depth estimates to compensate for physical attenuation and defocus blur, while graph-based temporal modeling captures dynamic gesture patterns across frames. This unified design ensures robust feature alignment and temporal coherence, enabling accurate classification of subtle gestures at distances up to 30 meters under low-resolution and noisy conditions \cite{godard2019digging}, which are common in hyper-range gesture recognition scenarios \cite{pla2018three, ren2011robust}.

To encourage the model to perform better at hyper-range distances, we introduce the Radiometric Spatio‐Temporal Depth Attenuation Loss (RSTDAL). This specialized loss function incorporates both Beer–Lambert attenuation and defocus weighting to adaptively adjust each training sample’s contribution based on its distance from the camera, thereby strengthening recognition accuracy for gestures performed at greater distances. The proposed DiG-Net trained using the RSTDAL loss yields a robust and generalizable recognition for dynamic gestures in both indoor and outdoor environments. The model addresses key challenges in hyper-range gesture recognition, such as the degradation of visual information due to reduced resolution and lighting variations \cite{gao2024challenges}. Furthermore, we introduce two novel metrics specifically designed to evaluate gesture recognition performance at long distances, including the stability of the recognition over time. Our approach is evaluated using these metrics, along with standard evaluation metrics.

To summarize, the key contributions of this work are:
\begin{itemize} 
\item We propose the novel Distance-aware Gesture Network (DiG-Net), which comprises a novel Depth-Conditioned Deformable Alignment (DADA) block alongside Spatio-Temporal Graph modules to robustly capture spatial deformations and temporal dynamics of gestures at hyper-range distances.
\item We introduce the Radiometric Spatio‐Temporal Depth Attenuation Loss (RSTDAL), a specialized loss function incorporating Beer–Lambert attenuation and defocus weighting to improve recognition robustness across varying distances, enabling effective hyper-range gesture recognition.
\item Unlike prior work, the proposed DiG-Net is the first to enable dynamic gesture recognition at hyper-range distances of up to 30 meters in both indoor and outdoor environments.
\item We provide a comprehensive evaluation of our model against state-of-the-art gesture recognition frameworks, demonstrating superior performance in challenging hyper-range scenarios. The evaluation is conducted using novel metrics specifically designed for hyper-range gesture recognition.

\item The trained models and datasets are publicly available to facilitate further research and development within the community\footnote{To be available upon acceptance for publication.}.
\end{itemize}
\hl{By extending reliable gesture recognition to hyper-range distances, this work aims to broaden the accessibility and usability of assistive robotic systems.}


\label{sec:introduction}

\section{Methods}

\subsection{Problem Formulation}
The primary objective of our work is for a robot to accurately recognize a human's dynamic hand gestures at distances of up to 30 meters in diverse environments. Given a set of $m$ gestures $O_1, O_2, \dots, O_m$, where each gesture can be either static or dynamic, the recognition task involves identifying the gesture performed in front of a simple RGB camera. The use of a standard RGB camera ensures that the dataset is applicable in real-world scenarios without the need for specialized hardware. Given an exhibited gesture $O_j$ captured by the camera in a video sequence $V_t=\{I_{t-n+1}, \dots, I_{t}\}$ of $n$ past frames at time $t$ ($I_t$ is a video frame taken at time $t$), we seek to maximize the conditional probability $P(O_j \mid V_t)$. Hence, our formulation aims to find $O_{j^*}$ acquired from the solution of optimization problem
\begin{equation}
    j^* = \underset{j}{\mathrm{argmax}} \; P(O_j \mid V_t), \quad j = 1, \dots, m~.
\end{equation}
Considering a sequence of past images enables the model to account for both the temporal and spatial dynamics of gestures, which is critical for differentiating between gestures that may appear similar when observed in a single frame.
\hl{Throughout this paper we consider a fixed vocabulary of $m=13$ gesture classes comprising eight dynamic gestures \{\textit{go-back}, \textit{go-up}, \textit{go-down}, \textit{move-right}, \textit{move-left}, \textit{turn-around}, \textit{beckoning}, \textit{follow-me}\}, four static gestures \{\textit{pointing}, \textit{thumbs-up}, \textit{thumbs-down}, \textit{stop}\}, and a \textit{null}} \hl{class indicating no intentional gesture. This vocabulary is used consistently for data collection, training, and evaluation; the dynamic gestures are illustrated in Figure}~\ref{fig:human_gestures}, \hl{with detailed definitions provided later in Section}~\ref{sec:gestures}.


\subsection{Data Collection}

\begin{figure}[htbp]
    \centering
    \includegraphics[width=\linewidth]{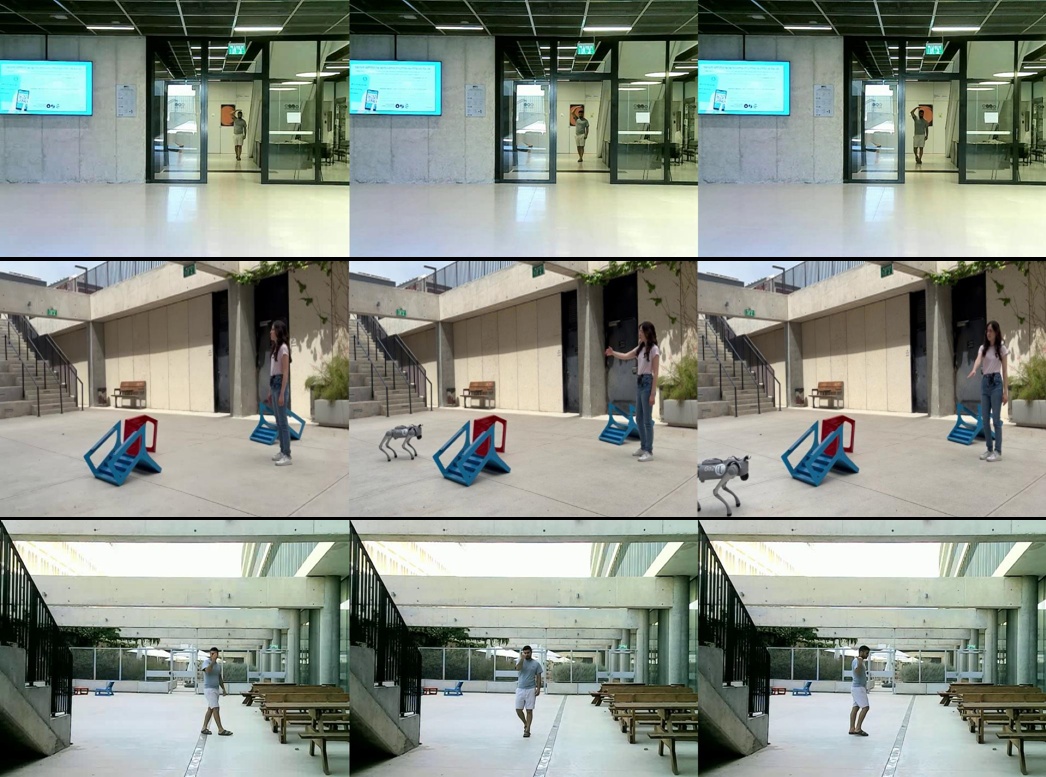}
    \caption{Example frames from the collected gesture dataset showing different users, gestures, and distances in indoor and outdoor environments.}
    \Description{Sample frames from the DiG-Net gesture dataset illustrating multiple users performing various dynamic gestures across different distances, lighting conditions, and backgrounds. 
    The dataset includes both indoor and outdoor scenes with diverse spatial arrangements to enhance robustness and generalization for hyper-range gesture recognition.}
    \label{fig:data_col}
\end{figure}

To train a gesture recognition model, a comprehensive dataset of hand gestures is required. A video $V_t$, i.e., a sequence of images captured by a simple RGB camera, is given, showing a user exhibiting a gesture $O_i$ at a distance $\rho_i\leq30~m$. The distance $\rho_i$ was measured manually using a standard measuring tape during data collection and annotated per sample. This value remains fixed for the entire video sequence and is used during training, regardless of cropping or other preprocessing steps. This yields a dataset $\mathcal{D} = \{(V_i, \rho_i, o_i)\}_{i=1}^{N}$ of $N$ labeled video sequences, where $o_i\in\{1,\ldots,m\}$ is the gesture index. Figure~\ref{fig:data_col} illustrates a diverse set of samples from the collected dataset, demonstrating the variety of environments, distances, and user poses used during data collection.

To enhance the dataset and improve the model's generalizability, data augmentation techniques are applied to simulate various real-world conditions. These techniques included random cropping, horizontal flipping, rotation, scaling, brightness and contrast adjustments, and synthetic noise addition. These are aimed at simulating different camera positions, lighting conditions, viewing angles, and environmental challenges. The augmented dataset $\tilde{\mathcal{D}} = \{(\tilde{V}_i, \rho_i, o_i)\}_{i=1}^{M}$ consists of both original and augmented video sequences, such that $M > N$. This enhanced dataset provides diverse training examples, thus improving the model's robustness and performance under real-world variability.

\subsection{User Study on Human Recognition of Long-Range Gestures}
\label{sec:User_Study}

To better understand how humans perceive gestures at extended distances, we conducted a user study that complements the technical evaluation of DiG-Net by examining the perceptual challenges humans face under similar conditions. The goal was to assess human recognition performance and subjective confidence when identifying static and dynamic gestures at varying distances, thereby providing a behavioral reference for interpreting the model’s capabilities. A total of ten participants (three female, seven male) took part in the experiment. \hl{The study followed a dense repeated-measures design: each participant evaluated approximately 50 randomly sampled trials spanning all gesture classes and the three distance conditions, yielding $\sim$500 individual human observations. This supports robust within-subject trend estimation across distance, while we note that $N{=}10$ limits population-level generalizability; larger and more diverse cohorts are left for future work.}

Each participant sat in front of a computer and was presented with both single-frame images and short video clips depicting human gestures recorded at three distances—short (5--10~m), medium (15--20~m), and long (25--30~m). All participants completed the study under controlled viewing conditions, using a full-HD monitor in a quiet, evenly lit indoor environment with minimal visual distractions. The gestures corresponded to those used in the DiG-Net dataset, including "stop", "come", "go back", "turn left", and "turn right". For each presented sample, participants were asked to (1) identify which gesture was shown, (2) press a key as soon as they recognized it (reaction time), and (3) rate their confidence in the identification on a 7-point Likert scale. The experiment revealed a clear distance-dependent effect on recognition accuracy and confidence. Participants achieved near-perfect recognition for both static and dynamic gestures at short range, while performance degraded substantially beyond 20~m, particularly for static frames where spatial details were lost. In contrast, dynamic gesture clips maintained relatively high recognition rates even at long distances, suggesting that motion cues play a key role in preserving interpretability. Qualitative feedback indicated that participants perceived dynamic gestures as "more natural and expressive", and several noted that motion helped them infer the actor’s intention even when fine-grained shape information was unclear. Confusion matrices further showed that errors primarily occurred between visually similar gestures, such as "stop" and "go back", especially at long range. These findings provide human-perception grounding for DiG-Net’s spatio-temporal design: while humans struggle with static, distance-degraded inputs, temporal motion dynamics enable more reliable interpretation. This supports the model’s emphasis on depth-compensated and motion-aware representations. Moreover, the results underscore the broader significance of long-range gestures as a natural, intuitive communication channel for guiding robots and establishing mutual understanding in assistive and collaborative scenarios. A quantitative summary of user study outcomes is provided in Table~\ref{tab:user_results} within the Evaluation section.

\subsection{Sequence Pre-processing}
\label{sec:preprocess}

A sequence $V_t$ contains $n$ images, some of which may be highly similar, leading to redundancy. Hence, the sequences in $\tilde{\mathcal{D}}$ are pre-processed to reduce the number of frames, resize them, and extract relevant features. Given a video $V_t$, the frames are reduced to $r < n$ representative frames using K-Means clustering. Each frame $\mathbf{I}_j\in V_t$ was processed through a pre-trained ResNet \cite{He} to extract a low-dimensional feature vector $\mathbf{\phi}_j \in \mathbb{R}^d$. All feature vectors were clustered into $r$ clusters, and the frame corresponding to the feature vector closest to each cluster's centroid was selected as the representative frame. In practice, a representative feature vector is chosen from cluster $C_i=\{\phi_1,\phi_2,\ldots\}$ according to
\begin{equation}
    \text{rep}(i) = \arg\min_{j} \|\mathbf{\phi}_j - \mathbf{C}_i\|_2,
\end{equation}
yielding $r$ representative frames $\{\mathbf{I}_{\text{rep}, 1}, \mathbf{I}_{\text{rep}, 2}, \ldots, \mathbf{I}_{\text{rep}, r}\}$. YOLOv3 \cite{Redmon} was then used to detect the user within each frame and crop the background, enhancing focus, particularly when the user was far from the camera. Full-body detection was chosen, as it provides more stable and consistent localization at hyper-range distances. In contrast, detecting small body parts like hands results in very limited regions, making the input noisier and less informative due to the loss of fine details. While YOLOv3 assists in isolating the user, it occasionally fails at long distances due to resolution loss. This emphasizes the importance of the model’s robustness in handling degraded inputs for reliable recognition. To address cases where the bounding box did not fully capture the human body, the bounding box was extended while maintaining a constant aspect ratio. Specifically, the pixel extension added around the bounding box was $\frac{b}{a}$, where $b$ is the diagonal length of the bounding box, and $a$ is a predefined user-to-image ratio parameter. The resulting cropped image was resized to $224 \times 224$ pixels to ensure uniformity across the dataset, maintaining a consistent size and aspect ratio.

To stabilize the training process, the frames were also normalized to have zero mean and unit variance. In addition, optical flow was computed between consecutive frames to capture motion dynamics, providing information on the direction and magnitude of motion, which is essential for distinguishing between gestures with similar spatial features but different temporal movements. The optical flow data was used as an additional input channel alongside the RGB frames, providing both spatial and temporal information to the model. Dataset $\tilde{\mathcal{D}}$ is updated to include the reduced and enhanced frames.

\subsection{DiG-Net Model Framework}
The proposed DiG-Net integrates DADA blocks with STG modules and Graph Transformer encoders to robustly handle spatial distortions and temporal dynamics of hand gestures at hyper-range distances. \hl{Our DADA modules build upon the deformable convolution framework introduced by Dai et al. }\cite{dai2017deformable}, \hl{extending it by conditioning the learned offsets on estimated depth and motion cues. This depth-aware adaptation enables the model to handle better spatial distortions and attenuation effects unique to hyper-range gesture recognition.} Individually, DADA blocks effectively compensate for physical attenuation and defocus blur, but cannot capture complex temporal dependencies. Conversely, STG modules and Graph Transformer encoders excel in modeling global spatio-temporal relationships but require well-aligned, noise-free inputs, making them sensitive to misalignment and distortions in low-resolution frames. DiG-Net addresses these complementary limitations by unifying depth-aware alignment, graph-based temporal reasoning, and global self-attention, thus enabling reliable recognition at distances of up to thirty meters. DiG-Net operates in a staged manner. Initially, input frames undergo successive DADA modules, which estimate motion-guided sampling offsets, perform ray-based feature warping, and correct for depth-related attenuation and defocus distortions. These depth-corrected features are then structured into a spatio-temporal graph, upon which the STG module executes spatial and temporal message passing to model local dynamics. Subsequently, Graph Transformer encoders apply multi-head self-attention across the graph nodes, effectively capturing long-range temporal dependencies and global contextual interactions. The self-attention mechanism of the Graph Transformer links early and late gesture phases, enhancing the importance of subtle deformation patterns identified through ray-sampling convolutions and resolving residual ambiguities inherent in distant, low-resolution inputs. Finally, the refined spatio-temporal features from the backbone network are concatenated into a tensor denoted by $\mathbf{Y}\in\mathbb{R}^{B\times c_{\mathrm{DADA}}\times T/4\times H/4\times W/4}$, where $B$ is the batch size, $c_{\mathrm{DADA}}$ represents the number of output channels after Graph Transformer processing, and $T$, $H$, and $W$ indicate the temporal length, spatial height, and spatial width of the input video sequence, respectively.

The initial feature map \(\mathbf{F}_0\) produced by the 3D convolutional stem is refined by the Motion–Depth Contextual OffsetNet, which predicts a dense offset field \(\Delta\mathbf{P}\) from the concatenation of \(\mathbf{F}_0\), the per‐pixel depth map \(z\), and optical‐flow channels \((u,v)\). This offset field guides the subsequent ray‐sampling convolution, which warps \(\mathbf{F}_0\) along the local motion direction. Specifically, we first define the sampled feature $\hat{\mathbf{F}}_0^{(k)} = \mathbf{F}_0\bigl(S_k(x,y,\tau)\bigr),$ 
where \(S_k(x,y,\tau)\) denotes the \(k\)-th sampling position along the normalized flow direction \(\hat f\), scaled by \(k / z(x,y,\tau)\), and is defined as
\begin{equation}
\label{eq:sampling}
S_k(x,y,\tau) = (x,y) + \frac{k}{z(x,y,\tau)}\,\hat f,
\quad
\hat f = \frac{(u,v)}{\|(u,v)\| + \epsilon}
\end{equation}
and \((x,y)\) are spatial pixel coordinates, \(\tau\) is the frame index, \(k\in[-K,K]\) is the sampling offset index, and \(\epsilon\) a small constant for numerical stability. Subsequently, the warped feature map is defined as  
\begin{equation}
\label{eq:warp}
F_{\mathrm{warp}}
= \sum_{k=-K}^{K}
  g_k\bigl(z(x,y,\tau),\,\|(u,v)\|\bigr)\,
  \hat{\mathbf{F}}_0^{(k)}
\end{equation}
where \(g_k\) is a learned weighting function of depth and flow magnitude \(\|(u,v)\|\). The warped features are then corrected for physical attenuation via  
\begin{equation}
\label{eq:attenuation}
\mathbf{F}_{\mathrm{corr}}
= F_{\mathrm{warp}} \,\exp\!\bigl(\eta\,z(x,y,\tau)\bigr)
\end{equation}
with \(\eta\) the Beer–Lambert attenuation coefficient \cite{swinehart1962beer}. After spatio‐temporal graph and Graph Transformer processing, the depth‐corrected features are aggregated over space and time:  
\begin{equation}
\label{eq:pooling}
\mathbf{h}_i = \frac{1}{T'H'W'}\sum_{\tau=1}^{T'}\sum_{x=1}^{H'}\sum_{y=1}^{W'}
\mathbf{F}_{\mathrm{corr}}^{(i)}(x,y, \tau),
\end{equation}
where \(T'=T/4\), \(H'=H/4\), and \(W'=W/4\) denote the pooled temporal and spatial dimensions. The resulting vector \(\mathbf{h}_i \in \mathbb{R}^{c_{\mathrm{DADA}}}\) is then fed into a fully connected layer followed by Softmax to produce the final probability distribution \(\hat{\mathbf{y}}_i \in \mathbb{R}^m\), where \(m\) is the total number of gesture classes.

To train the DiG-Net model, we propose the Radiometric Spatio-Temporal Depth Attenuation Loss (RSTDAL), a novel margin-based loss function tailored for robust gesture recognition at hyper-range distances. Unlike conventional loss functions that ignore distance-related degradations, RSTDAL incorporates depth-aware physical priors and motion dynamics to adaptively increase the decision margin for challenging inputs. Specifically, it penalizes the misclassification of gestures captured at far distances and with subtle motion, where attenuation and defocus effects severely degrade the signal. The loss function builds on an angular margin-softmax formulation, where the adaptive margin \( \mathcal{M}(\rho_i, \xi_i) \) is computed as:

\begin{equation}
\label{eq:rstdal_margin}
\mathcal{M}(\rho_i, \xi_i) = \gamma_1 (1 - e^{-\mu \rho_i}) + \gamma_2 Q + \gamma_3 \left(1 - e^{-\lambda \xi_i}\right),
\end{equation}

where Q = $\left(1 - \frac{1}{1 + (\rho_i/\rho_0)^2}\right)$ and \( \rho_i \) is the distance of the gesture, \( \xi_i \) is the average motion magnitude from the optical flow, and \( \mu, \rho_0, \lambda \) are learnable attenuation parameters. \hl{The intuition behind RSTDAL is to embed a physical prior directly into the decision geometry of the classifier. In hyper-range scenarios, signal quality degrades monotonically with distance, naturally reducing the separability between gesture classes. Standard losses often fail to prioritize these "hard" examples. RSTDAL addresses this by utilizing the Beer--Lambert attenuation term not to restore the signal, but to dynamically adjust the classification margin. By enforcing a larger margin for samples affected by high attenuation (distance) or low motion intensity, the objective function penalizes ambiguous embeddings in difficult conditions. This drives the network to learn distance-robust representations that rely on stable spatio-temporal dynamics rather than fine-grained features that are susceptible to degradation.}
The overall RSTDAL loss is then given by:

\begin{equation}
\label{eq:rstdal_loss}
\mathcal{L}_{\text{RSTDAL}} = -\frac{1}{B} \sum_{i=1}^{B} \log \frac{G}{G + \sum_{j\neq y_i} \exp\left(s \langle \mathbf{e}_i, \boldsymbol{\theta}_j \rangle \right)},
\end{equation}
where $G=\exp\left(s \left( \langle \mathbf{e}_i, \boldsymbol{\theta}_{y_i} \rangle - \mathcal{M}(\rho_i, \xi_i) \right) \right)$ and \( \mathbf{e}_i \) is the normalized embedding, \( \boldsymbol{\theta}_j \) are class prototypes, \( s \) is a scaling factor, and \( B \) is the batch size. By integrating distance, defocus, and motion terms into a unified margin, RSTDAL encourages better separation between classes under adverse visual conditions, significantly enhancing robustness across a wide range of distances and gesture speeds.

\label{sec:method}

\section{Model Evaluation}

In this section, we evaluate the proposed DiG-Net framework for recognizing dynamic gestures used to naturally guide a robot. All computations were conducted on a Linux Ubuntu 18.04 LTS system equipped with an Intel Xeon Gold 6230R CPU (20 cores at 2.1 GHz) and four NVIDIA GeForce RTX 2080TI GPUs, each with 11 GB of RAM. Hyperparameter tuning was performed using Ray-Tune \cite{Liaw2018} for all models.

\subsection{Gestures}
\label{sec:gestures}

\begin{figure}[htbp]
    \centering
    \includegraphics[width=\linewidth]{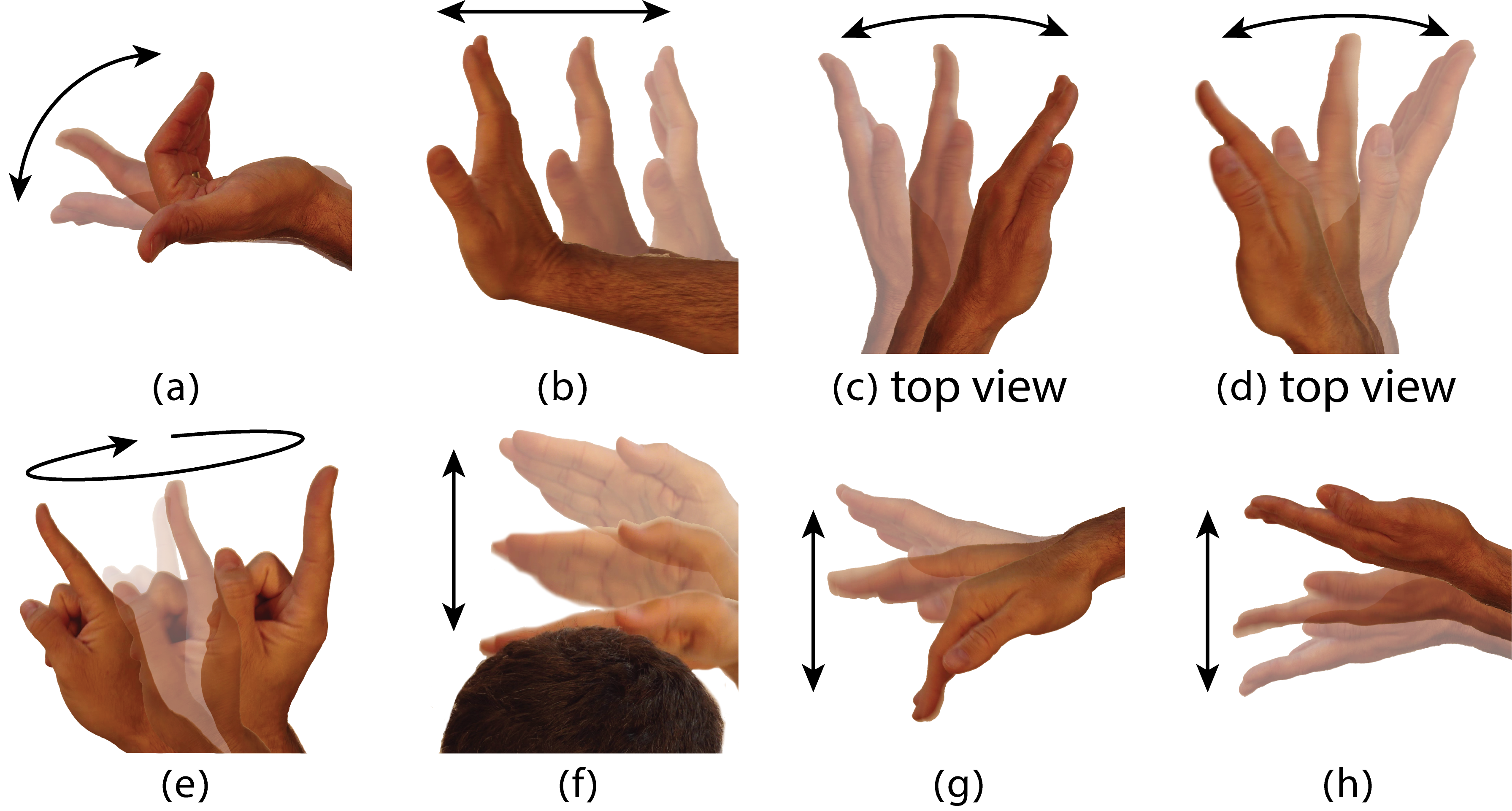}
    \caption{The eight dynamic gestures used in the analysis: (a) beckoning, (b) go-back, (c) move-right, (d) move-left, (e) turn-around, (f) follow-me, (g) go-down, and (h) go-up.}
    \Description{A visual illustration of the eight dynamic hand gestures in the DiG-Net dataset. 
    Each subimage shows a person performing one of the gestures: beckoning, go-back, move-right, move-left, turn-around, follow-me, go-down, and go-up. 
    The gestures are depicted as sequences of hand and arm movements, providing visual references for the actions analyzed in the study.}
    \label{fig:human_gestures}
\end{figure}

The evaluation focuses on $m=13$ distinct gesture classes. Of the 13, eight are dynamic, illustrated in Figure \ref{fig:human_gestures}, and include: \textit{go-back} with a forward and backward motion of an open hand, palm facing outward; \textit{go-up} with an upward motion of an open hand, palm facing upward; \textit{go-down} with a downward motion of an open hand, palm facing downward; \textit{move-right} with a horizontal sweeping motion of the open hand, palm facing right; \textit{move-left} with horizontal sweeping motion of the open hand, palm facing left; \textit{turn-around} with a circular motion of the index finger, pointing upwards; \textit{beckoning} where the palm is facing upward and the fingers are rhythmically flexed and extended; \textit{follow-me} where the open palm is repeatedly tapped on the head of the user. 

Another four static gestures are included: \textit{pointing}, \textit{thumbs-up}, \textit{thumbs-down}, and \textit{stop}. These gestures were chosen due to their mainstream usage and to challenge our model. That is, confusion between static and dynamic gestures may occur, such as in turn-around vs. pointing, go-back vs. stop, and go-up vs. beckoning. All gestures can be performed with either the left or right arm. The last class is the \textit{null}, which represents the absence of any exhibited gesture, in which the user can perform any unrelated activities. 

\subsection{Dataset}

Dataset $\mathcal{D}$ was collected by recording video samples of the $m=13$ gestures at a distance range of $\rho\in[2,30]$. Gestures were exhibited in diverse environments, including both indoor and outdoor settings, varying lighting, and diverse backgrounds. \hl{In addition, 16 different participants (11 males and 5 females, aged 25--44 years) contributed gesture data to ensure variability in gesture performance.} Each participant performed each gesture multiple times at different measured distances. The data collection process involved using a standard RGB camera with a resolution of $640 \times 480$ pixels taken at $21$ frames per second. Each gesture was manually annotated according to its class $o_i$ and the distance $\rho_i$ at which it was performed, resulting in labeled data suitable for supervised learning. This annotation allowed for a detailed analysis of model performance across varying distances, as well as the development of distance-specific recognition strategies. The collected dataset $\mathcal{D}$ yielded $N = 3,240$ video samples of hand gestures, each lasting 4 seconds with up to $n = 84$ frames. After feature extraction as described in Section \ref{sec:preprocess}, the number of frames was reduced to $k = 8$ per video. With further augmentation, dataset $\tilde{\mathcal{D}}$ includes $M=4,790$ samples. An additional test set of $K=458$ labeled and processed videos was recorded in distinct environments to evaluate the model's performance.

\subsection{Comparative Evaluation}

We now evaluate the proposed DiG-Net model compared to existing state-of-the-art gesture recognition frameworks. The DiG-Net model was trained with the Lion optimizer \cite{chen2023symbolic}, chosen for its efficiency and convergence stability, using an initial learning rate of 0.0037. The learning rate was gradually reduced using a cosine annealing schedule to ensure stable convergence. The model was trained for 100 epochs with a batch size of $B=16$, using the proposed RSTDAL loss function \eqref{eq:rstdal_loss}, which adaptively adjusts the angular margin based on the gesture distance $\rho_i \in [2, 30]$ and motion magnitude $\xi_i$ through the parameterized margin function $\mathcal{M}(\rho_i, \xi_i)$. The margin function used $\mu = 0.1$ and $\lambda = 0.2$ to control the saturation rates of the distance-driven and motion-driven margin terms, respectively, with a reference distance $\rho_0 = 16$ used in the non-linear margin term to center the response. The weights of the three margin components were set to $\gamma_1 = 0.4$, $\gamma_2 = 0.5$, and $\gamma_3 = 0.2$, respectively, to balance the influence of exponential decay over distance, the non-linear reference curve, and motion magnitude. These hyperparameters were optimized using RayTune \cite{Liaw2018} to balance the emphasis on long-range samples and ensure stable training. 

\hl{We analyzed the sensitivity of the RSTDAL hyperparameters. Specifically, $\mu$ and $\lambda$ control the saturation rate of the distance- and motion-driven margins. High $\lambda$ values cause the motion term $(1-e^{-\lambda \xi_i})$ to saturate rapidly, potentially over-penalizing subtle gestures, while low values render the motion cue negligible. Similarly, an excessively high $\mu$ causes the distance term $(1-e^{-\mu \rho_i})$ to dominate the gradient and bias optimization toward hyper-range samples, whereas a low $\mu$ fails to enforce sufficient separability at range. Additionally, $\rho_0$ sets the reference distance scale for the term $Q$, and $\gamma_{1,2,3}$ balances the contribution of the three margin components. The final values were selected via RayTune validation to balance validation performance across the full 3--30m range.}

To prevent overfitting, dropout was applied at the output of the Graph Transformer encoder, and L2 regularization was used during training. Early stopping was employed by monitoring the validation loss to determine the optimal stopping epoch. The DiG-Net model is compared to the: Swin Transformer \cite{Liu} which uses shifted windows for efficient spatial feature extraction; ViViT \cite{Arnab}, a Transformer for video analysis capturing spatial and temporal relationships; TimeSformer \cite{gberta_2021_ICML} which separates temporal and spatial attention; MViT \cite{fan2021multiscale}, a multiscale Transformer for spatiotemporal resolutions; I3D \cite{carreira2017quo}, a 3D convolutional network for spatiotemporal learning; and, X3D \cite{Feichtenhofer2}, an efficient 3D convolutional model balancing speed and accuracy. 

The models were trained and evaluated on the same datasets to ensure a fair comparison. We employ five key metrics: recognition success rate, Mean Average Precision (mAP), $F_1$ Score, Distance-Weighted Accuracy (DWA), and Gesture Stability Score (GSS). While the three former ones are common metrics, we propose to include the latter two for a comprehensive evaluation across distances and their consistency over time. The DWA emphasizes correct classifications of gestures performed at greater distances, reflecting the model's robustness. DWA is defined as:
\begin{equation}
\text{DWA} = \frac{1}{K} \sum_{i=1}^{K} \mathbb{I}(\tilde{o}_i = o_i)w_i ,
\end{equation}
where 
, $w_i = 1 + \beta \frac{\rho_i - \rho_{\text{min}}}{\rho_{\text{max}} - \rho_{\text{min}}}$, $\rho_i$ is the distance at which gesture $i$ was performed, $\rho_{\text{min}}=2~m$ and $\rho_{\text{max}}=30~m$ are the minimum and maximum distances, respectively, $\beta=1.6$ controls the weight for long distances, and $\mathbb{I}(\tilde{o}_i = o_i)$ equals to 1 if the predicted label $\tilde{o}_i$ matches the true label $o_i$, otherwise 0. While DWA evaluates the success rate with an emphasis on the farther cases, GSS measures the stability of predictions across the frames, ensuring consistent recognition. GSS is defined by
\begin{equation}
\text{GSS} = \frac{1}{K} \sum_{i=1}^{K} \left( \frac{1}{n_i} \sum_{j=n}^{n_i} \mathbb{I}(\tilde{o}_{i, j} = o_i) \right),
\end{equation}
where $n_i>n$ is the number of frames in video $V_i$, $\tilde{o}_{i, j}$ is the predicted label for sequence $\{I_{j-n+1},\ldots,I_j\}$ in video $V_i$, and $o_i$ is the true label for video $V_i$. A GSS value closer to one indicates a more stable prediction along the video. mAP provides an average measure of precision across all gesture classes, ensuring a balanced evaluation of the model's capability. $F_1$ Score is the harmonic mean of precision and recall, useful for evaluating models where both false positives and false negatives are impactful. These metrics provide a comprehensive evaluation of the DiG-Net model, focusing on accuracy, precision, and stability. Table \ref{tab:comparison} presents the comparative results after cross-validation, demonstrating the superior performance of the DiG-Net across all evaluation metrics, particularly in terms of recognition success rate.

\begin{table}[htbp]
\centering
\small
\setlength{\tabcolsep}{5pt}
\caption{Evaluation results for different dynamic gesture recognition models}
\label{tab:comparison}
\begin{threeparttable}
\begin{adjustbox}{width=\linewidth}
\begin{tabular}{lccccccc}
\toprule
\textbf{Model} & \textbf{Success rate (\%) $\uparrow$} & \textbf{DWA} $\uparrow$& \textbf{GSS} $\uparrow$& \textbf{$F_1$ score} $\uparrow$& \textbf{mAP (\%)} $\uparrow$& \textbf{Inference time (ms) $\downarrow$} \\
\midrule
Swin~\cite{Liu}           & 80.5 & 0.84 & 0.85 & 0.83 & 78.1 & 38 \\
ViViT~\cite{Arnab}        & 78.3 & 0.82 & 0.84 & 0.80 & 77.5 & 44 \\
TimeSformer~\cite{gberta_2021_ICML} & 83.4 & 0.85 & 0.87 & 0.85 & 81.3 & 49 \\
MViT~\cite{fan2021multiscale}       & 87.9 & 0.88 & 0.90 & 0.89 & 85.1 & 26 \\
I3D~\cite{carreira2017quo}          & 84.3 & 0.86 & 0.88 & 0.85 & 82.4 & 31 \\
X3D~\cite{Feichtenhofer2}           & 86.2 & 0.87 & 0.89 & 0.87 & 78.8 & 29 \\
CorrNet~\cite{hu2023continuous}     & 82.1 & 0.84 & 0.88 & 0.83 & 79.3 & 45 \\
GestFormer~\cite{garg2024gestformer} & 85.4 & 0.83 & 0.89 & 0.85 & 81.8 & 27 \\
\midrule
\rowcolor[HTML]{E0E0E0}
\textbf{DiG-Net} & \textbf{97.3} & \textbf{0.92} & \textbf{0.96} & \textbf{0.93} & \textbf{94.9} & \textbf{35} \\
\bottomrule
\end{tabular}
\end{adjustbox}
\end{threeparttable}
\end{table}

\subsection{DiG-Net analysis}

\begin{figure}[htbp]
\centering
\includegraphics[width=\linewidth]{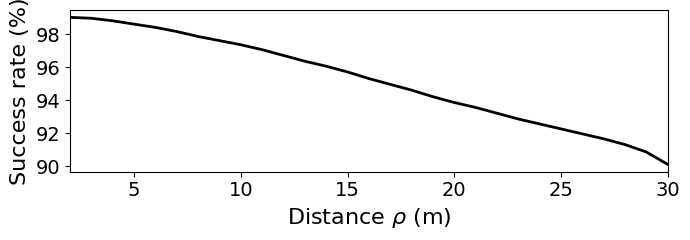}
\caption{Gesture recognition success rate of the DiG-Net model as a function of the user’s distance ($\rho$) from the camera. Performance gradually decreases at longer ranges due to lower image resolution and atmospheric effects.}
\Description{A line plot showing the recognition success rate of the DiG-Net model versus the user’s distance from the camera. 
Accuracy remains high at close and mid ranges and slowly declines as distance increases, demonstrating the model’s robustness up to approximately 30 meters. 
Axes are labeled with distance on the x-axis and success rate on the y-axis.}
\label{fig:performance_vs_distance}
\end{figure}

\begin{figure}[htbp]
    \centering
    \includegraphics[width=\linewidth]{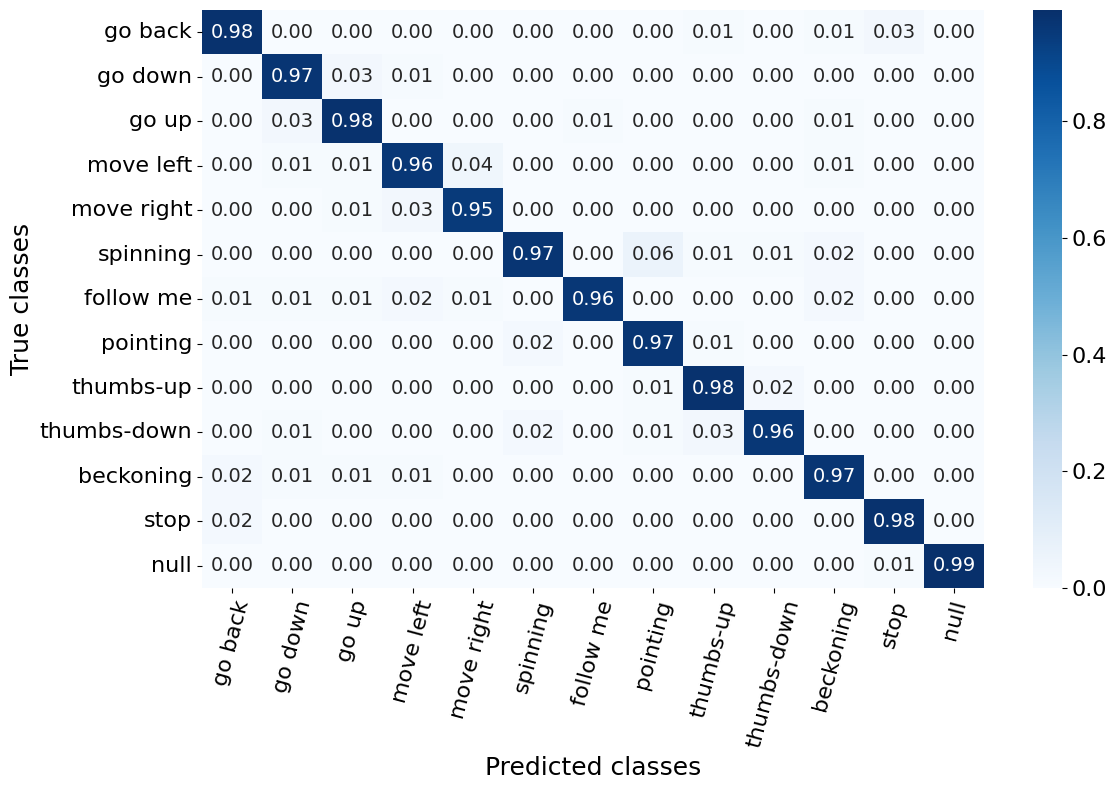}
    \caption{Confusion matrix for the gesture classification with the DiG-Net model across 13 gesture classes.}
    \Description{A heatmap representing the confusion matrix for 13 dynamic gesture classes using the DiG-Net model. 
    The x-axis represents predicted gesture classes, and the y-axis represents true classes. 
    Most values are concentrated along the diagonal, indicating strong classification accuracy, with minimal off-diagonal errors showing low confusion between gestures.}
    \label{fig:confmat}
\end{figure}

We further analyze the performance of the proposed DiG-Net model. Figure \ref{fig:performance_vs_distance} illustrates the gesture recognition success rate with respect to the distance $\rho$ between the user and the camera. The success rate gradually decreases but remains relatively high in the desired range of up to 30 meters. This demonstrates the robustness of the DiG-Net model in recognizing gestures at hyper-range, though performance diminishes as distance increases due to factors such as reduced resolution and increased visual noise. Figure \ref{fig:confmat} presents the confusion matrix for the DiG-Net model over the test data and for all 13 gesture classes.
While some gestures may be visually similar, the temporal context modeling enabled by the Graph Transformer encoder and spatio-temporal graph structure helps disambiguate subtle motion patterns, reducing the likelihood of misclassification.

\hl{To complement the quantitative evaluation, a qualitative analysis was conducted to better illustrate DiG-Net’s behavior under varying visual conditions. The confusion patterns in Figure}~\ref{fig:confmat}, \hl{together with the robustness results in Tables}~\ref{tab:combined_results}--\ref{tab:degradation_results}, \hl{reveal consistent trends in both success and failure regimes. DiG-Net demonstrates the highest reliability under hyper-range conditions, where fine-grained appearance cues vanish, and temporal information becomes the dominant discriminative factor. In these scenarios, the model maintains clear separability between gestures that are visually similar but differ in motion dynamics, such as \textit{go-back} and \textit{stop}, or \textit{turn-around} and \textit{pointing}. This behavior reflects the contribution of the depth-aware alignment and spatio-temporal reasoning modules, which jointly stabilize degraded inputs and preserve motion coherence. Failure cases primarily occur under extreme environmental degradation, consistent with the performance decline observed under severe clutter/interference and strong optical degradation conditions (Tables}~\ref{tab:noise_results} and~\ref{tab:degradation_results}). \hl{In such situations, low signal-to-noise ratios obscure fine motion details, occasionally causing confusion between gestures with similar overall trajectories when hand visibility is reduced. Overall, these qualitative findings align with the quantitative robustness analysis and highlight both the strengths and the remaining limitations of DiG-Net in real-world long-range conditions.}

\begin{figure}[htbp]
\centering
\includegraphics[width=\linewidth]{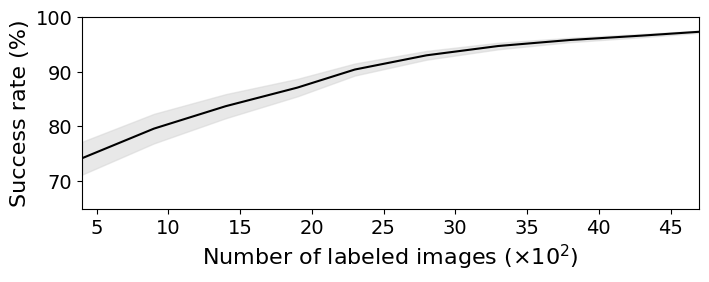}
\caption{Gesture recognition success rate of the DiG-Net model with regard to the number of labeled training images.}
\Description{A line plot showing the recognition success rate of the DiG-Net model versus the number of labeled training images. 
The curve rises sharply as more data are added and gradually plateaus, indicating that model accuracy saturates beyond roughly 25,000 labeled samples. 
Axes are labeled with the number of training images on the x-axis and success rate on the y-axis.}
\label{fig:performance_vs_data}
\end{figure}

To understand the impact of the amount of training data on the success rate of gesture recognition, we analyzed the DiG-Net model's performance with a varying number of labeled images. Figure \ref{fig:performance_vs_data} shows how the model's average success rate improves as more labeled images are utilized. For each number of labeled images, the DiG-Net model was trained 10 times with different parts of the dataset, yielding an average success rate. The success rate of the model increases significantly, reaching 97.3\% with the full dataset of 4,790 images. The gradual improvement with increased data demonstrates the model's ability to learn complex gesture patterns effectively as more labeled examples are introduced, but also indicates that beyond a certain point, adding more data does not yield a substantial gain. 

\begin{figure}[h]
\centering
\includegraphics[width=\linewidth]{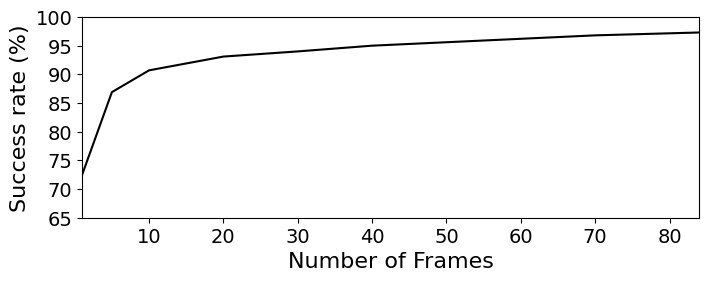}
\caption{Gesture recognition success rate of the DiG-Net concerning the number of frames $n$ in a video.}
\Description{A line chart showing the gesture recognition success rate of the DiG-Net model as a function of the number of frames per video sequence. 
The x-axis represents the number of frames ($n$), and the y-axis shows recognition success rate. 
The curve rises steadily as more frames are used, indicating improved temporal modeling, and plateaus once the sequence length exceeds around 40 frames.}
\label{fig:Success_Num_Frames}
\end{figure}

The above results were acquired with a video window length of up to $n=84$ frames. Hence, we now evaluate the performance of the DiG-Net with respect to the window length. The DiG-Net model was trained multiple times on video sequences $V_t$ of varying lengths $n$, and each trained model's performance was evaluated on test sequences of similar lengths. Figure \ref{fig:Success_Num_Frames} shows the recognition success rate of the DiG-Net with regard to the number of frames $n$ in a video sample. First, with only one image of the gesture (i.e., $n=1$), a poor recognition success rate of $72.5\%$ is achieved, due to the lack of dynamic information. As more frames are added to $V_t$, more dynamic information is included, and the recognition improves. A longer video encapsulates more dynamic information on the exhibited gesture, yielding better recognition accuracy. However, increasing $n$ may affect the time sensitivity and frequency of the recognition in real time. Nevertheless, increasing the sequence length beyond $n=84$ frames offers diminishing returns in terms of accuracy.

\begin{figure}[htbp]
\centering
\includegraphics[width=\linewidth]{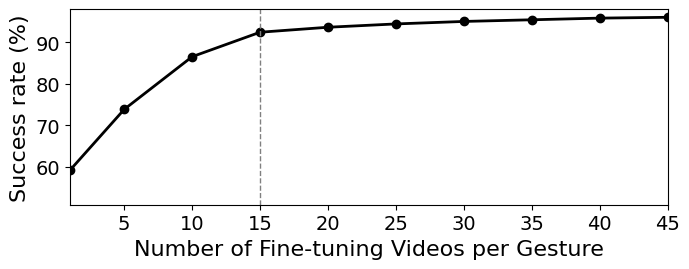}
\caption{Success rate of the DiG-Net model after fine-tuning to recognize five additional gestures, plotted against the number of video clips used for fine-tuning (each clip spans 84 frames).}
\label{fig:Finetune}
\end{figure}

We further evaluate the data requirements for adding new gestures to the DiG-Net classifier. Starting with a DiG-Net model trained to classify eight gestures with a success rate of 98.6\%, we examine the fine-tuning process needed to extend it to 13 gestures. Figure \ref{fig:Finetune} shows the success rate of the expanded model as a function of the number of video clips used for fine-tuning, with each clip containing 84 frames. The results indicate that high accuracy is achieved with just 15 video examples, with only marginal gains beyond that point. This highlights the model’s ability to generalize and adapt to new gesture categories with minimal additional data.

\begin{table}[h]
\centering
\small
\setlength{\tabcolsep}{6pt}
\caption{Ablation study results for components of the DiG-Net model. 
Each variant excludes one key module to evaluate its contribution to overall performance.}
\label{tab:Ablation}
\begin{threeparttable}
\begin{adjustbox}{width=\linewidth}
\begin{tabular}{lccccc}
\toprule
\textbf{Model variant} & \textbf{Success rate (\%)\,$\uparrow$} & \textbf{DWA\,$\uparrow$} & \textbf{GSS\,$\uparrow$} & \textbf{$F_1$ score\,$\uparrow$} & \textbf{mAP (\%)\,$\uparrow$} \\
\midrule
DiG-Net w/o DADA module            & 88.9 & 0.84 & 0.86 & 0.87 & 85.2 \\
DiG-Net w/o STG module             & 89.7 & 0.85 & 0.88 & 0.88 & 86.4 \\
DiG-Net w/o Graph Transformer      & 87.5 & 0.83 & 0.85 & 0.86 & 84.0 \\
DiG-Net w/o RSTDAL loss            & 90.1 & 0.86 & 0.89 & 0.89 & 87.1 \\
DiG-Net with short sequence        & 91.2 & 0.87 & 0.90 & 0.91 & 89.3 \\
\rowcolor[HTML]{E0E0E0}
\textbf{Full DiG-Net (proposed)}  & \textbf{97.3} & \textbf{0.92} & \textbf{0.96} & \textbf{0.93} & \textbf{94.9} \\
\bottomrule
\end{tabular}
\end{adjustbox}
\end{threeparttable}
\end{table}

An ablation study was also conducted to evaluate the contribution of each component of the DiG-Net model to its overall performance. The study involved training and evaluating the model while removing a single component at a time, including the DADA module, STG module, Graph Transformer encoder, and the RSTDAL loss. In addition, we examined the impact of training with shorter video sequences by reducing the input length to $n = 16$ frames. For the loss function ablation, we replaced the RSTDAL loss with the standard cross-entropy loss. The results are summarized in Table \ref{tab:Ablation}, demonstrating the importance of each component to the overall recognition accuracy, distance robustness, and temporal stability. The full DiG-Net model achieved the highest performance across all evaluation metrics, emphasizing the complementary contributions of depth-aware alignment, spatio-temporal graph reasoning, and global attention. Notably, removing the Graph Transformer or STG module resulted in a significant drop in accuracy and stability, while omitting the RSTDAL loss led to reduced emphasis on long-range gestures. Similarly, reducing the input sequence length limited the model's ability to capture temporal dynamics. These results highlight the necessity of each module for robust gesture recognition under hyper-range conditions.

\subsection{Human–Robot Evaluation and Comparative Insights}
To better contextualize the technical results of DiG-Net, we compared its quantitative performance with a user-based evaluation examining how humans perceive gestures at similar long-range conditions. This comparative perspective bridges model evaluation with human interpretability, an aspect central to HRI. The results of the user study (Section~\ref{sec:User_Study}) are summarized in Table~\ref{tab:user_results}. Ten participants (3 female, 7 male; age range 26–74; all with normal or corrected vision and no prior robotics experience) took part in the study. Participants rated their confidence on a 7-point Likert scale after each trial (1 = very uncertain, 7 = very confident), reflecting how sure they were about correctly identifying the gesture. Recognition accuracy and confidence decreased with distance, particularly for static gestures, whereas dynamic gestures remained comparatively easier to identify. Participants reported that temporal motion provided essential cues for understanding intent when spatial detail was limited. These findings mirror DiG-Net’s design philosophy, which explicitly leverages motion-aware and depth-compensated features to address similar perceptual degradations.
\begin{table}[htbp]
\centering
\small
\setlength{\tabcolsep}{5pt}
\caption{Summary of human recognition performance across gesture types and distances.}
\label{tab:user_results}
\begin{tabular}{lccc}
\toprule
\textbf{Condition} & \textbf{Accuracy (\%)} $\uparrow$& \textbf{Reaction Time (s)} $\downarrow$& \textbf{Confidence (1–7)} $\uparrow$\\
\midrule
Static (Short Range)   & 96.0 & 1.2 & 6.5 \\
Static (Mid Range)     & 82.0 & 1.9 & 5.3 \\
Static (Long Range)    & 68.0 & 2.5 & 4.7 \\
Dynamic (Short Range)  & 98.0 & 1.1 & 6.7 \\
Dynamic (Mid Range)    & 91.0 & 1.6 & 6.2 \\
Dynamic (Long Range)   & 84.0 & 1.8 & 5.8 \\
\bottomrule
\end{tabular}
\end{table}
Values in Table~\ref{tab:user_results} represent mean performance across all participants for each condition. While the study was exploratory and limited in scale, the reported values reflect consistent trends across participants. When comparing these behavioral findings to the model’s quantitative results, we observe that DiG-Net consistently outperforms human recognition at extended ranges, especially for static gestures where visual detail diminishes. For example, while participants achieved an average accuracy of 84\% for dynamic gestures at a long distance, DiG-Net maintained 94.9\% under the same conditions. This demonstrates that the model not only mirrors human motion sensitivity but also compensates for perceptual limitations that emerge with increasing range. Despite the modest sample size, the observed trends align with prior HRI perception studies and offer valuable qualitative insight into the interpretability of long-range gestures.


\subsection{Sequence Analysis and Varying Conditions}

To evaluate the robustness of the proposed DiG-Net model, we conducted a series of experiments involving complex gesture sequences and varying environmental conditions. In the first experiment, users performed complex gesture sequences consecutively, simulating real-world guidance scenarios for a robot. These sequences included various combinations of the 13 gesture classes. A total of 50 unique gesture sequences, each composed of 3 to 5 individual gestures, were used to ensure comprehensive testing.
Table \ref{tab:combined_results} presents the sequence accuracy, where a sequence is considered successful only if all gestures within it are correctly classified. The consistently high accuracy indicates that the DiG-Net model is capable of accurately interpreting and following multi-step gesture commands.

Additionally, we evaluated the model's recognition accuracy under varying lighting conditions, including indoor controlled lighting, outdoor bright sunlight, and overcast outdoor settings. The recognition success rates are seen in Table \ref{tab:combined_results}. The high accuracy in controlled indoor environments demonstrates the model's ability to recognize gestures under optimal conditions, while the slightly lower accuracy in outdoor environments reflects the challenges posed by changing lighting and environmental noise. Nevertheless, these results demonstrate the model's robustness in adapting to different environmental conditions, ensuring reliable performance in diverse real-world scenarios.

\begin{table}[h]
\centering
\small
\setlength{\tabcolsep}{8pt}
\caption{Results of Sequence Analysis and Varying Conditions}
\label{tab:combined_results}
\begin{threeparttable}
\begin{tabular}{lc}
\toprule
\textbf{Metric} & \textbf{Success rate (\%)\,$\uparrow$} \\
\midrule
Sequence accuracy            & 94.4 \\
Controlled lighting (indoor) & 96.8 \\
Bright sunlight (outdoor)    & 93.1 \\
Overcast (outdoor)           & 91.5 \\
\bottomrule
\end{tabular}
\end{threeparttable}
\end{table}

\subsection{Robustness and Inference Efficiency}

To further evaluate DiG-Net’s suitability for practical assistive robotics applications, we conducted experiments focused on robustness to environmental noise, real-time performance, and synthetic optical degradation. These tests simulate critical aspects of real-world assistive scenarios, ensuring the model maintains high reliability under challenging operational conditions.

Robustness to Environmental Noise:  
Assistive robotic systems frequently operate in dynamic and cluttered environments where irrelevant movements, background clutter, and sudden visual interferences occur frequently. To evaluate DiG-Net's robustness, we artificially injected background clutter, dynamic interference from moving objects, and abrupt lighting changes into existing recorded gesture sequences. Table~\ref{tab:noise_results} presents the recognition accuracy under different noise intensities.  

\begin{table}[htbp]
\centering
\small
\setlength{\tabcolsep}{8pt}
\caption{Recognition Accuracy under Environmental Noise Conditions}
\label{tab:noise_results}
\begin{threeparttable}
\begin{tabular}{lc}
\toprule
\textbf{Condition} & \textbf{Success rate (\%)\,$\uparrow$} \\
\midrule
Mild clutter/interference      & 95.8 \\
Moderate clutter/interference  & 93.5 \\
Severe clutter/interference    & 90.1 \\
\bottomrule
\end{tabular}
\end{threeparttable}
\end{table}

Real-Time Processing Simulation:
Real-time responsiveness is crucial in assistive scenarios, as delayed system reactions may lead to user frustration or incorrect robotic responses. We simulated real-time conditions by measuring inference speed in frames per second (FPS) on video sequences of varying length. Table~\ref{tab:realtime_results} summarizes the inference performance. Results confirm that the model supports real-time recognition even with longer input sequences. 

\begin{table}[htbp]
\centering
\small
\setlength{\tabcolsep}{8pt}
\caption{Inference Speed under Real-Time Simulation}

\label{tab:realtime_results}
\begin{threeparttable}
\begin{tabular}{lc}
\toprule
\textbf{Sequence length (frames)} & \textbf{Inference speed (FPS)\,$\downarrow$} \\
\midrule
8-frame sequence   & 28 \\
16-frame sequence  & 24 \\
32-frame sequence  & 19 \\
84-frame sequence  & 12 \\
\bottomrule
\end{tabular}
\end{threeparttable}
\end{table}

Synthetic Optical Degradation: 
To simulate realistic environmental challenges such as fog, defocus, or motion blur, we synthetically degraded gesture videos using Gaussian blur, motion blur, and synthetic fog effects at varying intensities. Table~\ref{tab:degradation_results} presents the model's accuracy under these visual degradations.

\hl{Deployment Considerations: To evaluate DiG-Net's suitability for practical assistive robotics deployment, we conducted real-time experiments on an NVIDIA Jetson Orin Nano embedded platform. The model processed gesture sequences of 8--32 frames at 15--25 frames per second, consistent with the simulated inference rates reported in Table~8. These results were achieved using the full-precision (FP32) model without any compression, quantization, pruning, or TensorRT acceleration, operating entirely within the platform's 8~GB memory limit. This confirms that DiG-Net supports real-time operation on resource-constrained embedded hardware commonly used in mobile robotics applications.}
\begin{table}[htbp]
\centering
\small
\setlength{\tabcolsep}{8pt}
\caption{Recognition Accuracy under Optical Degradation}
\label{tab:degradation_results}
\begin{threeparttable}
\begin{tabular}{lc}
\toprule
\textbf{Degradation level} & \textbf{Success rate (\%)\,$\uparrow$} \\
\midrule
Mild blur/fog      & 94.7 \\
Moderate blur/fog  & 91.2 \\
Severe blur/fog    & 88.3 \\
\bottomrule
\end{tabular}
\end{threeparttable}
\end{table}

\label{sec:Evaluation}

\section{Conclusions}

In this work, we proposed DiG-Net, a model for recognizing dynamic hand gestures at hyper-range distances of up to 30 meters using a monocular RGB camera. The model tackles key challenges in assistive robotics, including visual degradation, physical attenuation, and environmental noise in both indoor and outdoor settings. DiG-Net integrates DADA modules, spatio-temporal graph reasoning, and Graph Transformer encoders to capture spatial distortions and long-range temporal dependencies. We also introduced the RSTDAL loss, which adapts the decision margin based on distance and motion, improving recognition at extended ranges. Our experiments show that DiG-Net outperforms prior methods, achieving 97.3\% accuracy under challenging conditions.

The model significantly enhances the quality of life and independence for individuals with mobility impairments by enabling intuitive, long-range interaction with robotic assistants across various environments, including home healthcare, industrial safety, and emergency response. Future work will focus on expanding the assistive gesture vocabulary, integrating non-intentional body language cues, improving real-time responsiveness, and enhancing robustness in dynamic and crowded environments. \hl{We acknowledge that the user study sample size (N=10) is limited. Future work should include larger and more demographically diverse participant pools to strengthen the generalizability of the HRI findings.} These directions aim to bridge perceptual modeling with human-centered interaction design, contributing to the development of more transparent, adaptive, and socially aware assistive robotic systems.

\hl{In addition to the above contributions, we recognize several limitations that outline directions for future work. Our gesture dataset was collected from participants recruited in a single geographic region, and ethnicity and hand-size measurements were not explicitly balanced or recorded. In the accompanying user study, participants spanned a wider age range (26--74 years). At hyper-range distances, fine-grained hand morphology is strongly attenuated by resolution and imaging effects; accordingly, DiG-Net is designed to rely primarily on coarse spatiotemporal motion cues and full-body context rather than detailed hand-shape cues. Nevertheless, future work will aim to expand the dataset to include greater ethnic diversity to ensure broad generalization.}

\section*{Ethical Considerations}
All user study procedures followed ACM’s ethical guidelines for research involving human participants. All participants provided informed consent before participation, and all data were anonymized before analysis. The study did not involve any automated decision-making affecting individuals, and no personally identifiable information was collected or stored.

\section*{Conflict of Interest}
The authors declare that they have no known competing financial interests or personal relationships that could have influenced the work reported in this paper.

\section*{Funding}
This work was supported by the Israel Innovation Authority (grant No. 77857).

\section*{Acknowledgments}
The authors thank the participants and colleagues who assisted in the user study and provided valuable feedback.

\bibliographystyle{ACM-Reference-Format}
\bibliography{ref}

\end{document}